\documentclass{IEEEcsmag}

\usepackage[colorlinks,urlcolor=blue,linkcolor=blue,citecolor=blue]{hyperref}
\expandafter\def\expandafter\UrlBreaks\expandafter{\UrlBreaks\do\/\do\*\do\-\do\~\do\'\do\"\do\-}
\usepackage{upmath,color}
\usepackage{epsfig}
\usepackage{graphicx}
\usepackage{amsmath}
\usepackage{amssymb}
\usepackage{booktabs}
\usepackage{array}
\usepackage{multirow}
\usepackage{color}
\usepackage{colortbl}
\usepackage{framed}
\usepackage{bm}
\usepackage{bbm}
\usepackage{xspace}
\usepackage{cite}
\usepackage{xparse}
\usepackage{algorithm}
\usepackage{lipsum}
\usepackage{listings}
\usepackage{url}
\usepackage{duckuments}
\usepackage{algorithmic}

\makeatletter
\DeclareRobustCommand\onedot{\futurelet\@let@token\@onedot}
\def\@onedot{\ifx\@let@token.\else.\null\fi\xspace}

\def\eg{\emph{e.g}\onedot} 
\def\ie{\emph{i.e}\onedot}

\newcommand{\tit}[1]{\smallbreak\noindent\textbf{#1.}}
\newcommand{\tinytit}[1]{\noindent\textbf{#1.}}

\DeclareRobustCommand{\rchi}{{\mathpalette\irchi\relax}}
\newcommand{\irchi}[2]{\raisebox{\depth}{$#1\chi$}} 

\newcommand{\ours}{WF-Net\xspace}

\newcommand{\layername}[0]{WF-layer}
\newcommand{\layershortname}[0]{WF}
\newcommand{\layerfullname}[0]{Weight-Filtering (WF) layer}
\newcommand{\layermidname}[0]{Weight-Filtering\xspace}

\newcommand{\rev}[1]{\textcolor{black}{#1}}
\newcommand{\revv}[1]{\textcolor{black}{#1}}

\newcommand{\cout}[0]{C_{\text{out}}}
\newcommand{\cin}[0]{C_{\text{in}}}
\newcommand{\fout}[0]{F_{\text{out}}}
\newcommand{\fin}[0]{F_{\text{in}}}
\newcommand{\xyout}[1]{_{\text{out}}}
\newcommand{\xyin}[1]{_{\text{in}}}
\newcommand{\lossret}[0]{\lambda_0 \sum_{(x,y) \in \mathcal{D}_r} L_r\left(M(x), y\right)}
\newcommand{\lossunl}[0]{\lambda_1 \sum_{(x,y) \in \mathcal{D}_f} \frac{1}{L_f\left(M(x), y\right)}}

\newcommand{\regalpha}[0]{R\left(\hat{\alpha}\right)}

\jname{\textcolor{myblue}{IEEE Intelligent Systems}}
\pubyear{\the\year}

\begin{document}

\title{Multi-Class Unlearning\\for Image Classification via Weight Filtering}

\author{Samuele Poppi}
\affil{University of Modena and Reggio Emilia, Modena, 41125, Italy}

\author{Sara Sarto}
\affil{University of Modena and Reggio Emilia, Modena, 41125, Italy}

\author{Marcella Cornia}
\affil{University of Modena and Reggio Emilia, Reggio Emilia, 42121, Italy}

\author{{L}orenzo Baraldi}
\affil{University of Modena and Reggio Emilia, Modena, 41125, Italy}

\author{Rita Cucchiara}
\affil{University of Modena and Reggio Emilia, Modena, 41125, Italy}

\markboth{}{}

\begin{abstract}\looseness-1Machine Unlearning is an emerging paradigm for selectively removing the impact of training datapoints from a network. Unlike existing methods that target a limited subset or a single class, our framework unlearns all classes in a single round. We achieve this by modulating the network's components using memory matrices, enabling the network to demonstrate selective unlearning behavior for any class after training. By discovering weights that are specific to each class, our approach also recovers a representation of the classes which is explainable by design. We test the proposed framework on small- and medium-scale image classification datasets, with both convolution- and Transformer-based backbones, showcasing the potential for explainable solutions through unlearning.
\end{abstract}

\maketitle

\label{sec:intro}
\begin{figure*}[t]
\centerline{\includegraphics[width=0.7\linewidth]{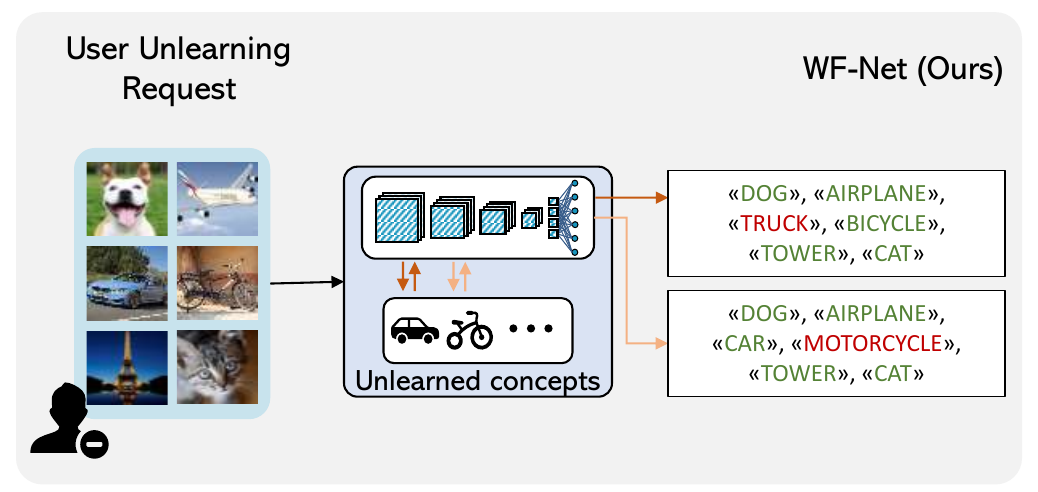}}
\vspace{-0.2cm}
\caption{The proposed single-round multi-class unlearning setting, which can unlearn \textit{any} class in a \textit{single} untraining round. \ours requires less computational resources and supports explainability by-design.}
\label{fig:firstpage}
\vspace*{-5pt}
\end{figure*}

\chapterinitial{Deep Learning models} and web datasets necessitate protecting user privacy and supporting the ``right to be forgotten"~\cite{dang2021right}. \revv{Among other techniques which can alter the internal knowledge representation of a model, like knowledge distillation, model pruning, or incremental learning, Machine Unlearning~\cite{nguyen2022survey} emerges as the sole technique which can delete private training data}. Machine unlearning approaches~\cite{7163042,izzo2021approximate,bourtoule2021machine}, indeed, remove traces left by specific datapoints from trained models. Their goal is to "untrain" the model, eliminating the impact of unwanted datapoints and reaching weights similar to those of models trained without such data. For instance, entire classes can be removed from classification networks~\cite{nguyen2020variational,neel2021descent,baumhauer2022machine}, which is useful in cases like face recognition. \revv{While existing machine unlearning approaches are limited to unlearning one or few classes jointly, our work advances by unlearning \textit{all} classes orthogonally and in a \textit{single fine-tuning round}. In particular, our approach can unlearn \textit{all} classes in an orthogonal fashion, ensuring the user a customizable, final model which can behave as a model unlearned on a single class of choice, imposed by the user at runtime. This  is obtained by exploiting} the mapping between inner network components (\eg, convolutional filters in a Convolutional Neural Network (CNN) or attentive projection in a Vision Transformer (ViT)) and output classes~\cite{wang2022hint}, and builds a \layermidname layer which can selectively turn on and off those inner components to accomplish the desired unlearning behavior on a class of choice (Figure~\ref{fig:firstpage}). This is achieved by encapsulating existing network operators, avoiding alterations to the network structure. By training the proposed \layermidname layers, we essentially uncover the connections between the inner network components and output classes. This not only enables effective unlearning but also provides an inherently interpretable representation\revv{, finally gaining explainability properties through unlearning}. Experimentally, we validate the proposed approach, named \ours, on small-scale and medium-scale image classification datasets and demonstrate its applicability to a variety of image classification architectures, including both CNNs and ViTs and to the more challenging case of unlearning without having access to the training set.

\smallskip
\noindent \textbf{Contributions.} 
To sum up, the contributions of this work are as follows:
\begin{itemize}
\item Our framework enables the simultaneous unlearning of multiple classes in a single round for an image classification network. This approach significantly reduces computational requirements during untraining and testing, offering greater flexibility compared to existing methods.
\item Our approach encapsulates inner network components like convolutional filters or attentive projections into \layermidname layers, which can selectively activate or deactivate these components to achieve the desired unlearning behavior.
\item Additionally, our method implicitly discovers the underlying relationships between convolutional filters or attentive projections and output classes and therefore allows to obtain a representation that can be employed for explainability purposes.
\item We conduct experiments on small and medium-scale image classification datasets, employing both CNN-based and ViT-based architectures. The results demonstrate the effectiveness of our proposed approach.
\end{itemize}

\section{RELATED WORK}
\label{sec:related}
Machine unlearning seeks to eliminate specific or sensitive data from pre-trained models, ideally without necessitating a full retraining process. In this realm, Cao~\emph{et al.}~\cite{7163042} were pioneers in tackling the machine unlearning problem within traditional machine learning algorithms. However, their approach exhibited limited advantages compared to retraining from scratch. 

\begin{figure*}[t]
\centerline{\includegraphics[width=\linewidth]{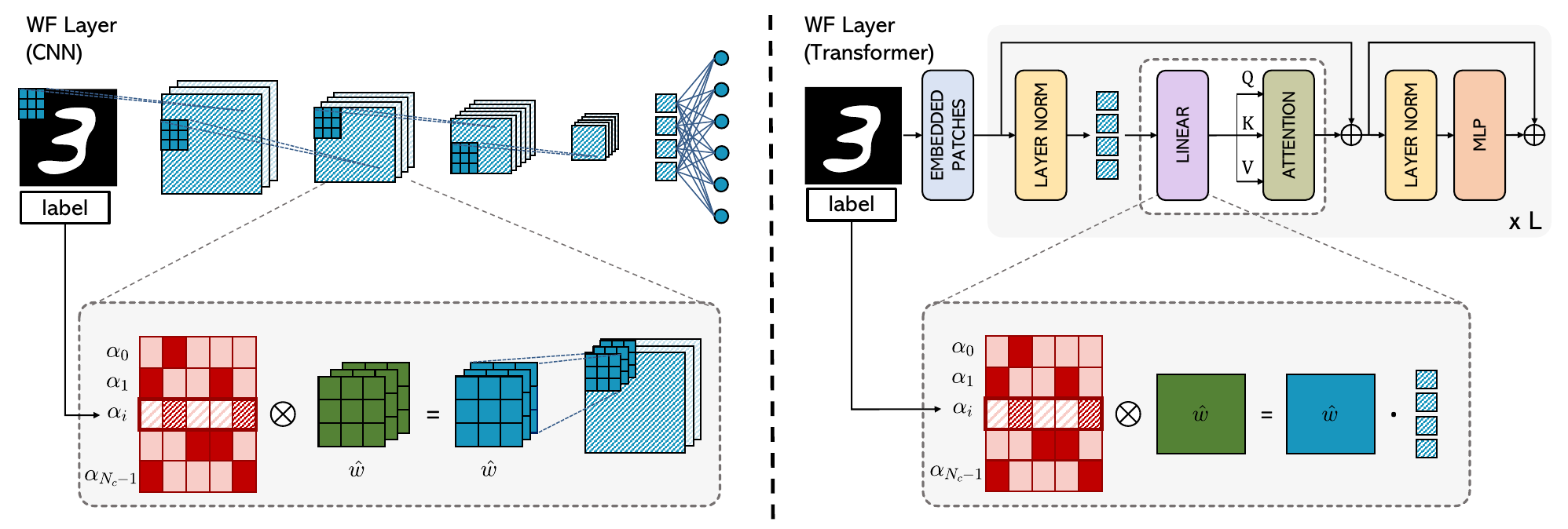}}
\vspace{-0.2cm}
\caption{Application of Weighted-Filter layers for single-shot multiple class unlearning on CNN-based and ViT-based architectures.}
\label{fig:model}
\vspace*{-5pt}
\end{figure*}

More recent efforts have been concentrated on incorporating unlearning into deep neural networks. While Izzo~\emph{et al.}~\cite{izzo2021approximate} introduced a technique for removing datapoints from linear models, other works involve a data grouping during training, enabling seamless unlearning by restricting the impact of datapoints on the model learning process~\cite{wu2020deltagrad}. Another approach, instead, is that of monitoring the impact of training datapoints on model parameters, thereby minimizing the amount of retraining required when a deletion request is received~\cite{bourtoule2021machine}. However, all the aforementioned methods lead to high storage costs. Another line of research proposed to employ efficient partitioning of training data. Among these works, Ginart~\emph{et al.}~\cite{ginart2019making} introduced a method for removing points from clustering by dividing the data into independent partitions, eventually using a distinct trained model for each partition.

Recent unlearning paradigms involve the removal of entire classes from a pre-trained model~\cite{nguyen2020variational,neel2021descent,baumhauer2022machine}. For example, Golatkar~\emph{et al.}~\cite{golatkar2020eternal} presented a technique for removing information from intermediate layers of deep learning networks via stochastic gradient descent, that can be also extended to the final activation of the model~\cite{golatkar2020forgetting}. \rev{Tarun~\emph{et al.}~\cite{tarun2021fast}, instead, introduced an effective class-wise unlearning framework in which an error-maximizing noise matrix for the forget class is learned and then used along with a subset of data the model should retain to update network parameters.}

Differently from previous works, we propose to unlearn all classes of a pre-trained model, orthogonally and in a single untraining stage, thus avoiding the need to save multiple unlearned models.

\section{PROPOSED METHOD}
\label{sec:method}
\subsection{Preliminaries}
We consider a set of input data $\mathcal{D} = \{ (x_i, y_i ) \}_{i=1}^n$ with $n$ number of samples, where $x_i$ is the $i$-th sample and $y_i$ is the corresponding label belonging to a set of classes $\mathcal{C}$. Traditional training aims at identifying a set of weights $\hat{w} \in \mathcal{W}$ via an iterative update rule $w_{t+1} = w_t - g(w, x')$, where $g(\cdot, x')$ is a stochastic gradient of a fixed loss function. Once the model has reached convergence and given a set of datapoints $\mathcal{D}_f$ drawn from the same distribution of $\mathcal{D}$, machine unlearning aims at identifying an update to $\hat{w}$ in the form
\begin{equation}
    \hat{w}_{t+1} = \hat{w}_{t} + g(z'_f), \text{with } \hat{w}_0 := \hat{w} \text{ and } z'_f \sim \mathcal{D}_f 
    \label{eq:sgd_unl}
\end{equation}
so that $\mathcal{D}_f$ is to be unlearned. In other words, the final set of weights of the unlearned model should be close to a model trained from scratch on $\mathcal{D} \setminus \mathcal{D}_f$. Depending on the use case, $\mathcal{D}_f$ can be conceived as containing a single datapoint (\ie, item removal), a group of data with similar features or labels (\ie, feature removal), or an entire class (class removal)~\cite{nguyen2022survey}.

\tit{Class removal} Forgetting data from single or multiple classes in a trained model presents challenges. Unlearning a class typically involves multiple passes over its datapoints, which is computationally costly compared to item or feature removal. Further, when unlearning different classes is required, the training procedure described in Eq.~\ref{eq:sgd_unl} must be repeated for each class as unlearning one class inevitably shifts $\hat{w}$ towards a configuration in which recovering a previously unlearnt class and then unlearning a new one becomes both cumbersome and inefficient. In the case where one wants a model to selectively unlearn all the $N_{c}$ classes on which it has been trained, this requires repeating the procedure $N_{c}$ times starting from $\hat{w}$ and store as many independent checkpoints.

\subsection{Single-shot multiple class unlearning}
Leveraging the known association between internal components and classes~\cite{wang2022hint}, our unlearning procedure spurs the model to recover these connections and store them in a learnable memory matrix. This is obtained by encapsulating internal network components in a \layerfullname, enabling orthogonal removal of each class in a single untraining step and resulting in a single final model checkpoint. At test time, one can simply select one of the $N_c$ classes and instruct the model to behave as if it was unlearned on the class of choice. An overview is shown in Figure~\ref{fig:model}.

\begin{algorithm}[t]
\caption{General forward step of a \layershortname~model}
\label{alg:forwardstep}
\begin{algorithmic}
\REQUIRE a \layershortname~Model $M$ comprising a sequence of layers $\mathcal{L}$, an input tensor $X_{\text{in}}$, a label $Y_{\text{unl}}$
\ENSURE a predicted Label $Y_{\text{out}} \gets M(X_{\text{in}}, Y_{\text{unl}})$
\STATE Initialize $X_{0} \gets X_{\text{in}}$
\FOR{$i \in \mathbb{N}_{\left|\mathcal{L}\right|}$}
    \STATE $l \gets \mathcal{L}[i]$
    \STATE $\hat{w}_l \gets \bm{\alpha}_l[Y_{\text{unl}}] \odot \hat{w}_l$
    \STATE $X_{i+1} \gets l(X_{i}, \hat{w}_l) $
\ENDFOR
\RETURN $\text{argmax}(X_{\left|\mathcal{L}\right|})$
\end{algorithmic}
\end{algorithm}

\tit{Intuition and single-class scenario} \rev{We introduce \ours with the description of the case in which we restrict to unlearning a single class only.}
Given a model $M:\mathcal{X}\rightarrow\mathcal{Y}$, a layer $l$ and its set of learned parameters $\hat{w}_l$, we encapsulate $l$ into our \layername. The \layershortname~then incorporates the original layer and a learnable memory vector $\bm{\alpha}_l$ having a shape of $1 \times \mathcal{K}$, where $\mathcal{K}$ is the cardinality of the original learnable weights. \rev{The memory vector is meant to mask the learned weights $\hat{w}_l$ and filter the information reaching the following deeper layers. The parameters of $\bm{\alpha}_l$ are real numbers, which are squeezed into the interval $[0,1]$ by applying a sigmoid function.}

\rev{Concretely, $\bm{\alpha}_l$ modulates the contributions of the inner weights of the layer, thus obtaining a modified layer $\hat{l}(\cdot)$ which behaves as
\begin{equation}
    \hat{l}(\cdot) = l(\cdot, \alpha_l \odot \hat{w}_l).
\end{equation}
The modified layer equals the original layer when all the elements of $\bm{\alpha}_l$ are set to $1$. Analogously, $\bm{\alpha}_l$ can act as an unlearned layer, by dropping some of the $\alpha$ scores to $0$.}
During this procedure, the model's original parameters are kept frozen (only $\bm{\alpha}_l$ is actually trained). Recalling the dependency between inner components and classes, a good unlearning procedure should then drop the minimum, yet optimal, set of $\alpha$ scores to $0$, to completely forget the objective, unlearning class.

\tinytit{Multi-class scenario} While the above-mentioned approach can provide an alternative to classic unlearning for the single-class case, it does not solve the time and storage issues arising when dealing with more classes is required. However, since our procedure exclusively relies on learnable memory vectors, it can be easily extended to the multi-class scenario. \rev{By extending $\bm{\alpha}_l$ to $N_{\text{c}}\times\mathcal{K}$ matrices, each row vector can memorize class-specific information. At forward time, given an unlearning class label, it will be sufficient to index the corresponding row in $\bm{\alpha}_l$ and apply the layer (see Algorithm~\ref{alg:forwardstep}).} Finally, coupled with a suitable untraining protocol, this allows performing a single procedure to unlearn all classes, orthogonally.
\rev{Finally, it is noteworthy to mention that our approach could be extended to multi-label datasets as well, by selecting a random ground-truth label as row selectors for the $\alpha_l$ matrices when computing the unlearning loss.}

\subsection{Applications to image architectures}
To prove its generalization capabilities, we apply our \layermidname~to both CNN- and Transformer-based architectures.

\tit{CNNs} \rev{In CNN-based architectures, we mask both the convolution kernels and biases, using the output channels as the masking granularity. Given a convolutional layer $l$ with kernels $\hat{w}^c_l\in\mathbb{R}^{\cout \times \cin \times kH \times kW}$ and biases $\hat{w}^b_l\in \mathbb{R}^{\cout}$, we apply two weighting matrices over each output channel, $\bm{\alpha}_l^c \in \mathbb{R}^{N_{\text{classes}}\times \cout}$ over $\hat{w}^c_l$ and $\bm{\alpha}_l^b \in \mathbb{R}^{N_{\text{classes}}\times \cout}$ over the biases.}
This results in masking the $\cout$ convolutive filters, with shape $\cin \times kH \times kW$ and having $\cout$ activations, weighted by a selected row of $\bm{\alpha}_l^c$. As well, each of the $\cout$ biases is weighted by a selected row of $\bm{\alpha}_l^b$. 

\tit{Vision Transformers} In the case of a Transformer-based network, we instead mask the weights $\hat{w}^c_l \in \mathbb{R}^{\fin \times \fout}$ and the biases $\hat{w}_l^b \in \mathbb{R}^{\fout}$ of the linear projection generating queries, keys, and values of each attentive layer $l$.

\subsection{Training}
Taking inspiration from previous works~\cite{chundawat2022can}, we employ two classification loss functions, namely an \textit{unlearning} loss $L_f$ and a \textit{retaining} loss $L_r$. The first one encourages the model to forget a specific class $c_f$, while the second one measures the capacity to retain the information about all the other classes ($c_r \in \mathcal{C} \setminus c_f$). We implement both of them as cross-entropy loss functions.

\rev{To avoid the shortcomings of the negative gradient approach~\cite{golatkar2020eternal}, obtained by inverting the sign of the gradient, we suggest minimizing the following objective, in which we employ the reciprocal of the forget loss:}
\begin{align}
    L = \lossret + \nonumber
    \\ \lossunl.
    \label{eq:3waysumcls}
\end{align}
Consequently, minimizing Eq~ \ref{eq:3waysumcls} towards zero, implies $L_f(\cdot)$ to be maximized, and $L_r(\cdot)$ to be minimized.

\tit{Regularization}
To ensure that only a few elements of $\bm{\alpha}_l$ are dropped to zero during untraining, we add a regularization term, enforcing the elements of $\bm{\alpha}_l$ to be nearly all active, except for a closed number of its parameters, which will serve as gates. We, therefore, add a regularizer $\regalpha$ to the final loss function.

\tinytit{Label expansion} To achieve simultaneous untraining of all classes, we avoid physically splitting the dataset $\mathcal{D}$ into two retaining and unlearning subsets $\mathcal{D}_r$ and $\mathcal{D}_f$, as each sample of the dataset could be employed to unlearn its designated class as well as to retain all the others. Instead, given a randomly sampled mini-batch (of length $B$), we split it into two halves, to obtain $B/2$ images and labels which are employed for unlearning and $B/2$ which are employed for retaining. Samples from the first half are employed, together with their ground-truth label, as row selectors for the $\bm{\alpha}_l$ matrices, and used to compute the unlearning loss. The second half, instead, is paired with randomly selected rows of $\bm{\alpha}_l$, and employed to compute the retaining loss with its ground-truth label.

It shall be noted that during optimization, every image of class $c$ will unlearn over the same row of $\bm{\alpha}_l$ (\eg, class $0$ will unlearn over the $0$-th row). When retaining, every class will randomly retain on one of the other rows, thanks to the random selection in $\bm{\alpha}_l$. This implies that every class will have full impact over its row while having considerably less impact for the retaining rows. To compensate for this disparity, we replicate the ``retaining'' component of the mini-batch $\rchi$ times, pairing it with differently sampled labels, so that every retaining image has a $\rchi$-times greater impact. 
As a result, the retaining loss has a shape of $(\rchi, B/2)$, while the untraining loss has a shape of $(B/2,)$. The last step is to average both losses.

\section{EXPERIMENTAL EVALUATION}
\label{sec:experiments}
\subsection{Datasets}
We perform experiments on three well-known image classification datasets: MNIST, CIFAR-10, and ImageNet. MNIST consists of 60,000 training and 10,000 test images, each corresponding to one of the 10 classes representing handwritten digits. CIFAR-10 contains 60,000 images across 10 classes, with 6,000 images per class, split into training (50,000) and validation (10,000) sets. ImageNet comprises 1.28 million training images and 50,000 validation images, each associated with one of the 1,000 classes of the dataset. Evaluation on ImageNet involves a randomly selected subset of the validation set with 5,000 images, featuring five images for each class.

\subsection{Evaluation Metrics}
We evaluate \ours in terms of evaluation metrics that measure the unlearning capabilities of the model as well as metrics that estimate its level of explainability.

\tinytit{Accuracy on retain and forget sets} We employ accuracy scores to measure the correctness of predictions for both retain and forget sets. Ideally, the accuracy on the retain set should be close to the one of the original model before unlearning, while the accuracy on the forget set should be close to zero or equal to the accuracy of a model re-trained without samples of the forget class.

\tinytit{\rev{Relearn Time (RT)}} \rev{It measures the number of epochs taken by the model to regain full accuracy on a class from the forget set. Following previous works~\cite{tarun2021fast}, we train the model on 500 random samples from the training set in
each epoch, stopping training if the model does not reach the original accuracy after 100 epochs. In our experiments, we report the relearn time averaged over all classes of the dataset, showing in parentheses the number of classes the model was unable to relearn after 100 epochs.}

\tinytit{Zero Retrain Forgetting (ZRF) score} It estimates the randomness of the unlearned model by comparing it with a randomly initialized network. In particular, this metric compares the output probabilities of the two models using a Jensen-Shannon divergence ($JS$). Following previous literature~\cite{chundawat2022can}, we compute the ZRF score as follows:
\begin{equation}
    \text{ZRF} = 1-\frac{1}{N_f}\sum_{i=0}^{N_f} JS\left(M(x_i), M^*(x_i) \right),
\end{equation}
where $M$ is the unlearned model, $M^*$ is its randomly initialized version, $x_i$ is the $i$-th sample from the forget set, and $N_f$ is the number of elements of the forget set. The final score lies between 0 and 1, where a score near 1 corresponds to a completely random behavior of $M$. 

\begin{table*}[t]
\vspace*{4pt}
\caption{Machine Unlearning results on MNIST, CIFAR-10 and ImageNet-1k, in comparison with original and retrained models. \rev{The $\dagger$ marker indicates single-class unlearning methods.}}
\label{tab:results}
\centering
\setlength{\tabcolsep}{.35em}
\resizebox{\linewidth}{!}{
\begin{tabular}{cclc ccc>{\color{black}}c c ccc>{\color{black}}c c ccc}
\toprule
& & & & \multicolumn{4}{c}{\textbf{MNIST}} & & \multicolumn{4}{c}{\textbf{CIFAR-10}}  & & \multicolumn{3}{c}{\textbf{ImageNet-1k}} \\
\cmidrule{5-8} \cmidrule{10-13} \cmidrule{15-17}
 & & & & Acc$_r$ $\uparrow$ & Acc$_f$ $\downarrow$ & ZRF $\uparrow$ & RT $\uparrow$ & & Acc$_r$ $\uparrow$ & Acc$_f$ $\downarrow$ & ZRF $\uparrow$ & RT $\uparrow$ & & Acc$_r$ $\uparrow$ & Acc$_f$ $\downarrow$ & ZRF $\uparrow$ \\
\midrule
& & Original Model & & 99.6 & 99.6 & 48.0 & - & & 93.0 & 93.0 & 48.3 & - & & 71.2 & 71.3 & 0.35 \\
& & Retrained Model & & 99.4 & 0.0 & 48.7 & 74.5 (5) & & 89.9 & 0.0 & 50.1 & 65.3 (5) & & - & - & - \\
\cmidrule{3-17}
& & \rev{Random Labels$^\dagger$} & & \rev{99.6} & \rev{0.0} & \rev{53.3} & \rev{40.3 (0)} & & \rev{88.7} & \rev{0.0} & \rev{52.5} & \rev{56.5 (5)} & & \rev{-} & \rev{-} & \rev{-} \\
& & \rev{Negative Gradients$^\dagger$} & & \rev{99.4} & \rev{0.0} & \rev{48.3} & \rev{24.6 (0)} & & \rev{89.4} & \rev{0.0} & \rev{51.6} & \rev{76.7 (7)} & & \rev{-} & \rev{-} & \rev{-} \\
\cmidrule{3-17}
\cellcolor{white}\multirow{-5}{*}{\textbf{VGG-16}} & \cellcolor{white} & \textbf{\ours} & & 73.2 & 0.0 & 79.2 & 27.6 (1) & & 80.2 & 18.3 & 57.4 & 54.2 (5) & & 64.1 & 1.89 & 0.53 \\
\midrule
& & Original Model & & 99.6 & 99.6 & 47.0 & - & & 93.9 & 94.0 & 48.0 & - & & 70.5 & 70.3 & 0.35 \\
& & Retrained Model & & 99.4 & 0.0 & 48.7 & 43.6 (0) & & 90.5 & 0.0 & 51.4 & $>$ 100 (10) & & - & - & - \\
\cmidrule{3-17}
& & \rev{Random Labels$^\dagger$} & & \rev{99.4} & \rev{0.0} & \rev{57.1} & \rev{15.7 (0)} & & \rev{90.6} & \rev{0.0} &\rev{59.4} & \rev{21.6 (0)} & & \rev{-} & \rev{-} & \rev{-} \\
& & \rev{Negative Gradients$^\dagger$} & & \rev{99.2} & \rev{0.0} & \rev{52.1} & \rev{20.1 (0)} & & \rev{92.4} & \rev{0.0} & \rev{55.2} & \rev{31.3 (3)} & & \rev{-} & \rev{-} & \rev{-} \\
\cmidrule{3-17}
\cellcolor{white}\multirow{-5}{*}{\textbf{ResNet-18}} & \cellcolor{white} & \textbf{\ours} & & 94.0 & 9.68 & 63.1 & 22.7 (0) & & 79.7 & 9.25 & 63.9 & 94.3 (9) & & 64.4 & 1.47 & 0.40 \\
\midrule
& & Original Model & & 98.9 & 98.9 & 47.2 & - & & 78.0 & 78.0 & 49.8 & - & & 75.6 & 75.5 & 0.34 \\
& & Retrained Model & & 99.0 & 0.0 & 50.2 & 71.4 (7) & & 71.2 & 0.0 & 67.6 & 52.2 (4) & & - & - & - \\
\cmidrule{3-17}
& & \rev{Random Labels$^\dagger$} & & \rev{98.3} & \rev{0.0} & \rev{48.7} & \rev{81.9 (8)} & & \rev{73.2} & \rev{0.0} & \rev{57.7} & \rev{83.5 (8)} & & \rev{-} & \rev{-} & \rev{-} \\
& & \rev{Negative Gradients$^\dagger$} & & \rev{98.4} & \rev{0.0} & \rev{54.6} & \rev{71.6 (7)} & & \rev{74.4} & \rev{0.0} & \rev{50.1} & \rev{81.7 (8)} & & \rev{-} & \rev{-} & \rev{-} \\
\cmidrule{3-17}
\cellcolor{white}\multirow{-5}{*}{\textbf{ViT-T}} & \cellcolor{white} & \textbf{\ours} & & 93.5 & 0.0 & 47.4 & $>$ 100 (10) & & 73.5 & 0.0 & 59.8 & $>$ 100 (10) & & 68.0 & 2.51 & 0.44 \\
\midrule
& & Original Model & & 99.0 & 98.9 & 47.1 & - & & 85.2 & 85.2 & 53.6 & - & & 82.3 & 82.2 & 0.34 \\
& & Retrained Model & & 99.0 & 0.0 & 49.5 & 91.5 (8) & & 75.5 & 0.0 & 61.3 & 51.1 (4) & & - & - & - \\
\cmidrule{3-17}
& & \rev{Random Labels$^\dagger$} & & \rev{99.5} & \rev{0.0} & \rev{50.4} & \rev{74.1 (7)} & & \rev{74.8} & \rev{0.0} & \rev{50.8} & \rev{75.0 (6)} & & \rev{-} & \rev{-} & \rev{-} \\
& & \rev{Negative Gradients$^\dagger$} & & \rev{99.6} & \rev{0.0} & \rev{52.3} & \rev{44.1 (4)} & & \rev{75.5} & \rev{0.0} & \rev{51.7} & \rev{61.7 (5)} & & \rev{-} & \rev{-} & \rev{-} \\
\cmidrule{3-17}
\cellcolor{white}\multirow{-5}{*}{\textbf{ViT-S}} & \cellcolor{white} & \textbf{\ours} & & 94.0 & 0.0 & 48.0 & 91.0 (9) & & 74.7 & 0.0 & 59.7 & 81.1 (8) & & 69.2 & 9.46 & 0.45 \\
\bottomrule
\end{tabular}
}
\vspace*{-5pt}
\end{table*}

\tinytit{Insertion score} The proposed metric measures explainability map goodness~\cite{petsiuk2018rise}. In a few words, it computes the rise in the confidence score as the image pixels are iteratively reactivated, by relevance, from a baseline (usually a $0$ tensor). In our case, $\bm{\alpha}_l$ matrices contain key information on forgotten classes. Unlike the original implementation, at each iteration, we reactivate a percentage of elements in each $\bm{\alpha}_l$ sorted by relevancy, evaluate the network's confidence score, and normalize it by baseline model's confidence score, \ie~the corresponding model before untraining.
Eventually, the average AUC score is taken. The closer the score is to $1$, the more significant the features selected by the $\bm{\alpha}_l$ matrices are.

\tinytit{Deletion score} It represents the counterpart of the insertion score, as it measures the drop in score as the image pixels are zeroed, by relevance, at each iteration until the initial picture becomes a baseline tensor. In our case, we again maneuver our $\bm{\alpha}$ scores to show the correlation between their values and the information about the unlearned classes. To compute the score, we start from a non-unlearned \ours and, iteratively, drop a percentage of them to $0$, from most relevant to least relevant ones. Each iteration records the normalized confidence score of the \ours for unlearned classes, relative to the baseline model's confidence score. The average AUC of the resulting curve represents the final score.

\subsection{Implementation and Training Details}
\label{sub:impl-det}
We conduct our investigation with various backbones with a model size compatible with both middle and medium-scale datasets, namely VGG-16, ResNet-18, and ViT in its Tiny (ViT-T) and Small (ViT-S) versions.

During untraining, we initialize all $\bm{\alpha}_l$ to $3$, so that $\sigma(\bm{\alpha}) \simeq 1$ and have a non-zero gradient during the initial stages of learning. We employ a fixed (un)learning rate of $100$, which allows us to escape from the minimum reached during the pre-training, and a fixed mini-batch size of $128$. In the CNNs case, we employ $\lambda_0 = 1$, $\lambda_1 = 10$, $\lambda_2 = 1$. In the case of ViT we raise $\lambda_1$ to $100$. The label expansion factor $\rchi = 3$ in all the cases, and eventually, we early stop the untraining procedure with a patience of $10$, evaluating the validation loss $5$ times per epoch.

\subsection{Experimental Results}

\tinytit{Unlearning performance}
We start by assessing the unlearning performance of our approach. 
Table~\ref{tab:results} shows results on MNIST, CIFAR-10, and ImageNet training sets. We compare with the performances of a model trained on the full training set from scratch (termed as ``original model''), and with a collection of models trained without each of the classes. We train these models only for MNIST and CIFAR-10, due to the higher number of classes in the ImageNet dataset. The average performance of these models is termed ``retrained model''. \rev{Additionally, we implement two single-class unlearning methods, namely \textit{random labels}~\cite{hayase2020selective} and \textit{negative gradients}~\cite{golatkar2020eternal}. In the former, we fine-tune the model using randomly assigned labels for samples from the forget set, while in the latter, the model is fine-tuned on the forget set using negative gradients (\ie, fine-tuned in the direction of gradient ascent). It is worth noting that the unlearning phase of these two methods is performed separately for each class. For this reason, we do not report the results of these methods on the ImageNet dataset.}

As it can be observed, \ours exhibits a proper unlearning behavior across all classes and models. In particular, we notice that on MNIST and CIFAR-10 it showcases a 0.0 accuracy on the forget sets when trained with VGG-16, ViT-S and ViT-T, underlying that the proposed strategy is effective in removing the knowledge of a class of choice, even without directly updating the model weights and while managing the unlearning of more classes concurrently. In terms of accuracy on the retain set, instead, we observe a restrained loss with respect to the performances of both the full model and of the retrained model. For example, on the ImageNet-1k dataset the CNN-based versions of \ours only lose a few accuracy points on the retain set compared to the accuracy of the original model (\eg, from 70.5 to 64.4 with ResNet-18)\footnote{\rev{We noticed similar results also for larger backbones like ResNet-50 and ResNet-101, with a retaining accuracy respectively of 72.6 and 74.4 and a forget accuracy equal to 0.}}. With Transformer-based architectures, instead, the loss in terms of retain accuracy is slightly more evident (\eg, from 82.3 to 69.2 with ViT-S). When considering the ZRF score, computed on the unlearned classes, we can notice that the performance reached by our approach is always higher than that of the original model and generally in-line or higher than that of the retrained model, underlying again that our approach can properly unlearn.

\rev{Furthermore, our solution performs well also against single-class unlearning methods, although they require separate training for each class to be unlearned. In particular, while in terms of accuracy scores \ours performs slightly worse than random labels and negative gradients, it obtains significantly superior results in terms of ZRF score in most cases further confirming the unlearning capabilities of our solution. Considering instead the relearn time, our approach takes a comparable or higher number of epochs to relearn unlearned classes than all other models especially when a ViT-based model is used as backbone.}

\tit{Insertion and deletion scores}
In Table~\ref{tab:explainability} we report the insertion and deletion scores, according to the four model architectures. Modifying the $\alpha$ values has an impact on the scores given to images belonging to unlearned classes, and not on images belonging to other classes. This highlights that the proposed approach can identify filters and attention weights which are responsible of the identification of each class. 

\begin{table}[t]
\vspace*{4pt}
\caption{Insertion (Ins) and Deletion (Del) scores of the \ours model.}
\label{tab:explainability}
\centering
\setlength{\tabcolsep}{.32em}
\resizebox{\linewidth}{!}{
\begin{tabular}{lc cc c cc c cc}
\toprule
& & \multicolumn{2}{c}{\textbf{MNIST}} & & \multicolumn{2}{c}{\textbf{CIFAR-10}} & & \multicolumn{2}{c}{\textbf{ImageNet}} \\
\cmidrule{3-4} \cmidrule{6-7} \cmidrule{9-10}
& &  Ins $\uparrow$ & Del $\downarrow$ & &  Ins $\uparrow$ & Del $\downarrow$ & &  Ins $\uparrow$ & Del $\downarrow$ \\
\midrule
\textbf{VGG-16} & & 0.78 & 0.09 & & 0.79  & 0.10 & & 0.73 & 0.05  \\
\textbf{ResNet-18} & & 0.79 & 0.05 & & 0.69 & 0.06 & & 0.75 & 0.04 \\
\midrule
\textbf{ViT-T} & & 0.90 & 0.07 & & 0.88 & 0.05 & & 0.60 & 0.04  \\
\textbf{ViT-S} & & 0.90 & 0.08 & & 0.88 & 0.14 & & 0.75 & 0.04 \\
\bottomrule
\end{tabular}
}
\vspace*{-5pt}
\end{table}

\begin{figure}[t]
\centering
\setlength{\tabcolsep}{.01em}
\resizebox{\linewidth}{!}{
\includegraphics[height=.2\linewidth]{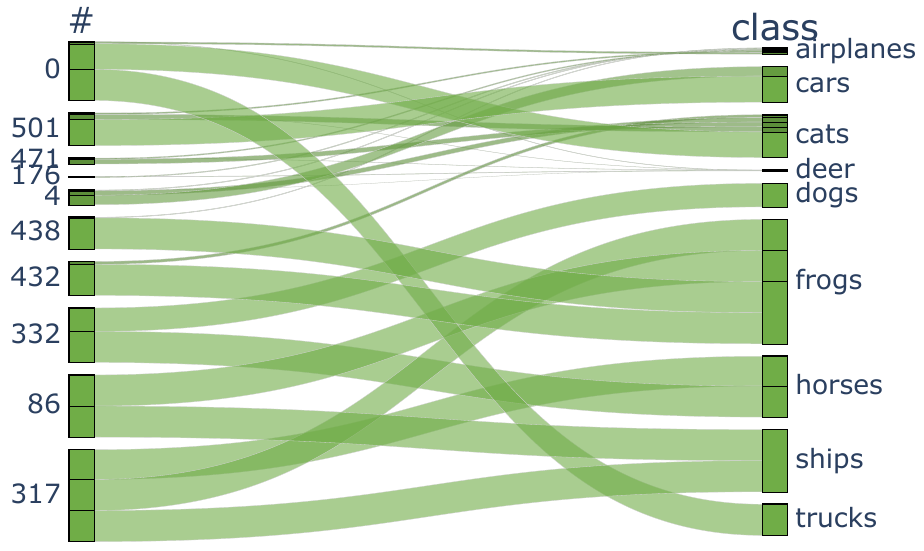}
}
\vspace{-0.3cm}
\caption{Relationships highlighted by \ours between the weights of a VGG-16 layer and the CIFAR-10 classes.}
\label{fig:visualizations}
\vspace*{-5pt}
\end{figure}

\begin{figure*}[t]
\centering
\small
\setlength{\tabcolsep}{.01em}
\resizebox{\linewidth}{!}{
\begin{tabular}{ccc}
 First Layer & Intermediate Layer & Last Layer \\
\includegraphics[height=.2\linewidth]{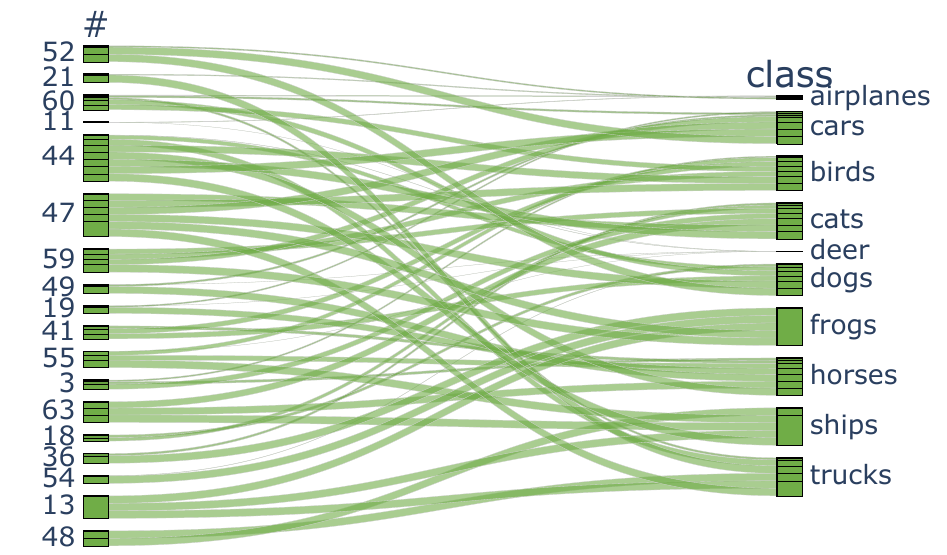} & 
\includegraphics[height=.2\linewidth]{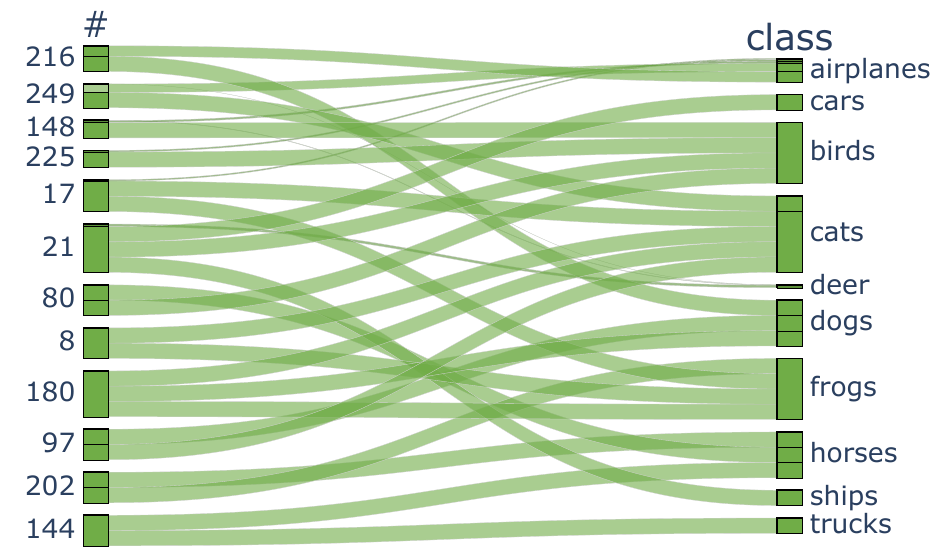} & 
\includegraphics[height=.2\linewidth]{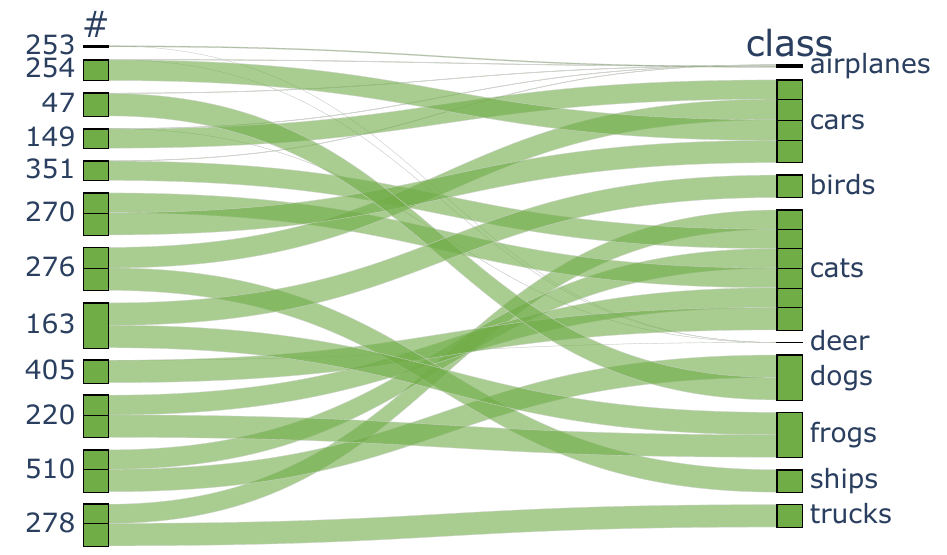} \\ 

\includegraphics[height=.2\linewidth]{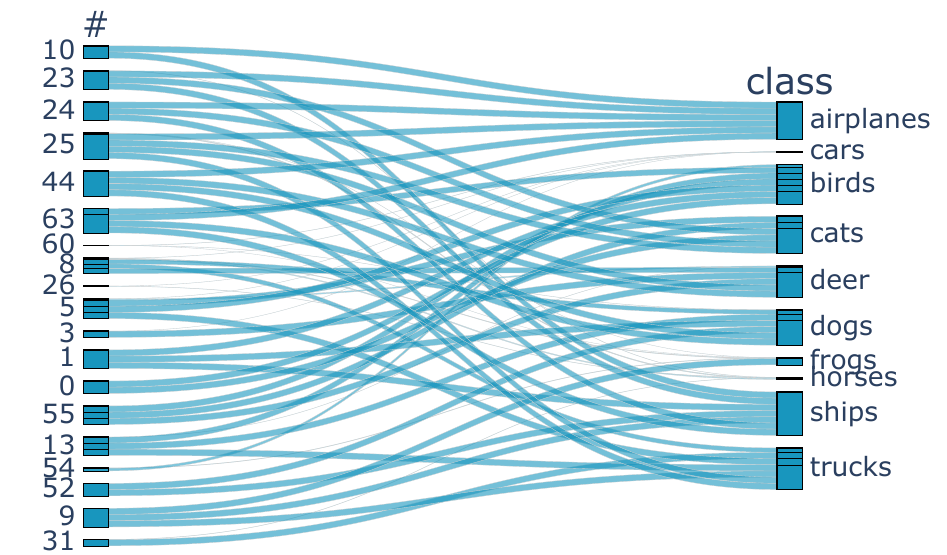} & 
\includegraphics[height=.2\linewidth]{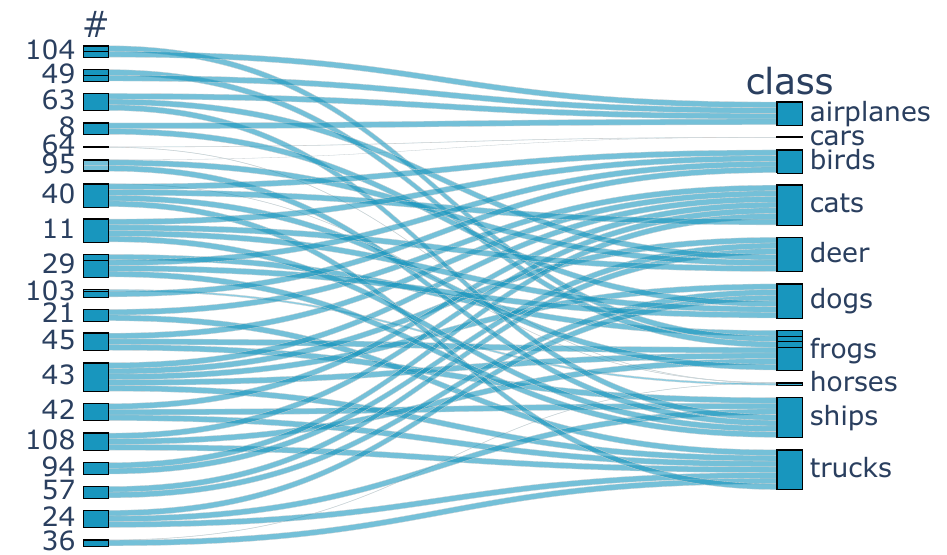} & 
\includegraphics[height=.2\linewidth]{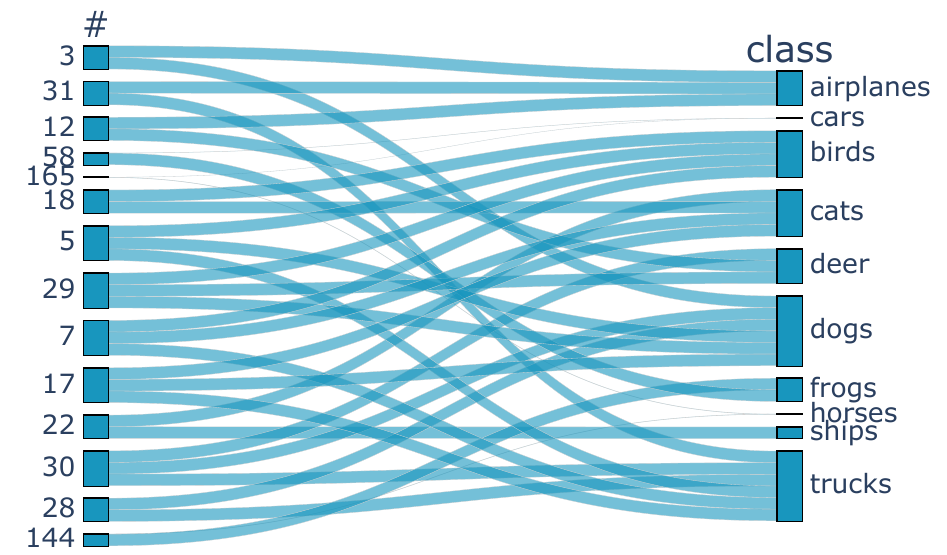} \\ 

\includegraphics[height=.2\linewidth]{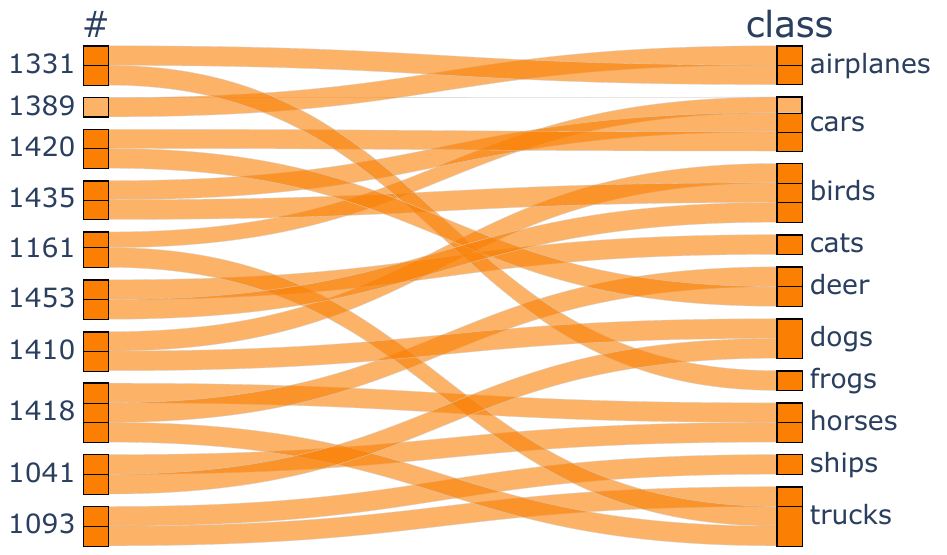} & 
\includegraphics[height=.2\linewidth]{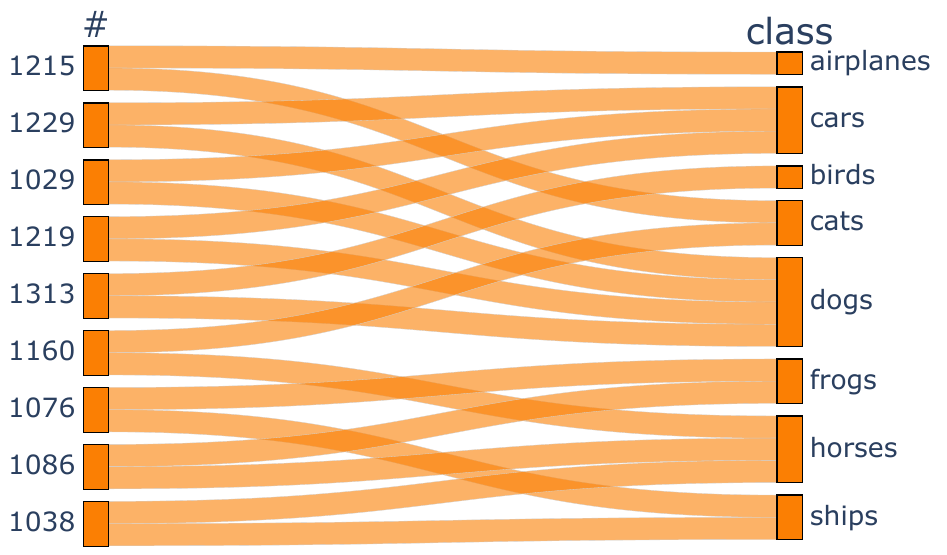} & 
\includegraphics[height=.2\linewidth]{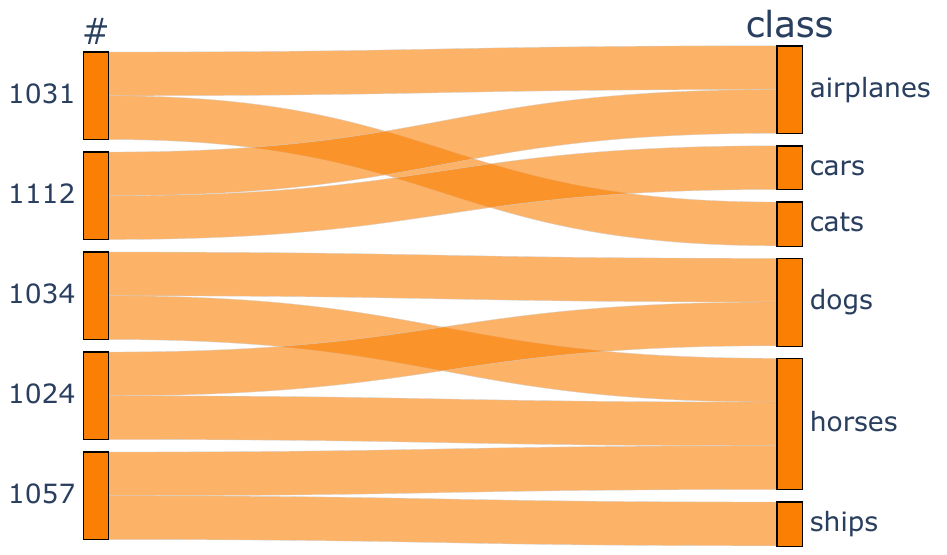} \\ 
\end{tabular}
}
\vspace{-0.1cm}
\caption{Application of Weighted-Filter layers for single-round multiple class unlearning on the CIFAR-10 dataset on different layers, using VGG-16 (first row), ResNet-18 (second row), and ViT-T (third row).}
\label{fig:visualizations_supp}
\end{figure*}

\tit{Filter visualizations}
Finally, in Figure~\ref{fig:visualizations} we qualitatively visualize the association between filters (or attentive projections) and output classes, as discovered by the $\alpha$ values after unlearning, for randomly selected layers. We proceed to select the top-10 filters for each class and remove those which do not appear in at least two classes, in order to spot relationships between classes. As can be noticed, the proposed approach discovers proper relationships between filters and classes. For example, we can notice how the filter \#332 contributes to two different classes (\ie, \textit{dogs} and \textit{horses}), while the filter \#3 contributes to the classes \textit{cats}, \textit{deer}, and \textit{dogs}. These patterns can help to better understand the model behavior and can highlight the underlying relationships within the network. Additional visualizations are shown in Figure~\ref{fig:visualizations_supp}.

\section{CONCLUSION}
\label{sec:conclusion}
We presented a novel approach for single-round multi-class unlearning. Given an image classification network, our approach can unlearn all classes simultaneously, in a single unlearning round. After training, the resulting network can be instructed to behave as if it has been untrained on any of the classes. Experimentally, we validated its effectiveness on classical image datasets and across different architectures, also showing that our approach can discover relationships between convolutional filters or attentive projections and output classes. 

\section{ACKNOWLEDGMENTS}
This work has been supported by the the EU Horizon project ``ELIAS - European Lighthouse of AI for Sustainability'' (No. 101120237), funded by the European Commission.

\def\refname{REFERENCES}

\begin{IEEEbiography}{Samuele Poppi} is a PhD student of the Italian National PhD Program in AI at the University of Modena and Reggio Emilia, 41125 Modena, Italy. His research interests include explainable AI and machine unlearning. Contact him at samuele.poppi@unimore.it.
\end{IEEEbiography}

\begin{IEEEbiography}{Sara Sarto} is currently pursuing the PhD degree in ICT at the University of Modena and Reggio Emilia, 41125 Modena, Italy. Her research interests include image captioning, cross-modal retrieval, and attentive models. Contact her at sara.sarto@unimore.it.
\end{IEEEbiography}

\begin{IEEEbiography}{Marcella Cornia} is a tenure track assistant professor with the University of Modena and Reggio Emilia, 42121 Reggio Emilia, Italy. She has coauthored more than 70 publications mainly on vision-and-language integration, attentive and saliency models. Contact her at marcella.cornia@unimore.it.
\end{IEEEbiography}

\begin{IEEEbiography}{Lorenzo Baraldi} is a tenure track assistant professor with the University of Modena and Reggio Emilia, 41125 Modena, Italy. He has coauthored more than 100 publications, mainly working on video understanding, deep learning, and multimedia. Contact him at lorenzo.baraldi@unimore.it.
\end{IEEEbiography}

\begin{IEEEbiography}{Rita Cucchiara} is a full professor at the University of Modena and Reggio Emilia, 41125 Modena, Italy, where she is the director of the Artificial Intelligence Research and Innovation Center, coordinating the AImageLab research lab and the Modena ELLIS Unit. Her research interests are related to computer vision, pattern recognition, and multimedia. Contact her at rita.cucchiara@unimore.it.
\end{IEEEbiography}

\newpage

\end{document}